\documentclass{article}
\usepackage{arxiv}
\pdfoutput=1
\usepackage[ruled,vlined]{algorithm2e}
\usepackage{multirow}
\usepackage{graphicx}
\usepackage[font=small]{caption}
\usepackage{subcaption}

\newcommand{\showDOI}[1]{\unskip}
\newcommand{\wmaxi}{\mbox{$w_i^{max}$}}
\newcommand{\Nmaxi}{\mbox{$N_i^{max}$}}
\date{}

\begin{document}
\title{Probabilistic spike propagation for FPGA implementation of spiking neural networks}

\author{Abinand Nallathambi\\
IIT Madras\\
ee16s032@ee.iitm.ac.in
\and Nitin Chandrachoodan\\
IIT Madras\\
nitin@ee.iitm.ac.in
}
\maketitle

\begin{abstract}
Evaluation of spiking neural networks requires fetching a large number of synaptic weights to update postsynaptic neurons. This limits parallelism and becomes a bottleneck for hardware.

We present an approach for spike propagation based on a probabilistic interpretation of weights, thus reducing memory accesses and updates. We study the effects of introducing randomness into the spike processing, and show on benchmark networks that this can be done with minimal impact on the recognition accuracy.

We present an architecture and the trade-offs in accuracy on fully connected and convolutional networks for the MNIST and CIFAR10 datasets on the Xilinx Zynq platform.
\end{abstract}

\section{Introduction}
\label{sec:intro}

Spiking neural networks are often referred to as the third generation of neural network models \cite{Maass1997NetworksModels}, and have several characteristics that make them attractive from the viewpoint of hardware design.  
They follow an event-driven model of computation, where the work done (hence energy consumed) can be made proportional to the number of spike events, and do not require the arrays of multiply-accumulate (MAC) operations that characterize conventional artificial neural networks (including convolutional networks, referred to here as ANNs)
This makes ANNs well-suited to parallel implementation on architectures such as GPUs.  

In contrast, spiking neural networks may require other types of computations to determine whether a neuron is to fire or not.  
Hardware architectures for spiking networks (eg. \cite{Akopyan2015TrueNorth:Chip, Furber2014TheProject, Davies2018Loihi:Learning, Neil2014MinitaurAccelerator}) therefore differ considerably from those for regular ANNs, and focus more on features that enable efficient event-driven computation.
This usually requires the network to be trained specifically for the target architecture (due to restrictions on permitted connections or weights), and it is not efficient to take a network trained for one architecture and directly run it on another.

Despite being event driven, spiking networks still require a large number of memory accesses, primarily for two purposes \cite{Neil2014MinitaurAccelerator}: determining the recipient neurons of a spike, and fetching weights of the corresponding synapses. 
Recent data (eg. \cite{Han2016EIE:Network}) indicates that fetching data from memory (especially off-chip) is much more expensive in energy than arithmetic computations.
For reasonably sized networks, the neuron weight and index information becomes too much to store on-chip -- the resulting off-chip memory accesses thus end up dominating the energy consumed for the computation, and can also lead to increased latency.

The regular view of a spiking network is that if a presynaptic neuron spikes, it increases the membrane potential of the postsynaptic neuron by an amount equal to the weight of the synapse.  
Alternatively, similar to the ideas in \cite{Seung2003LearningTransmission, Kasabov2010ToModel}, we could view weight as a measure of how likely it is that a spike will propagate across a synapse.

In this paper, we present a probabilistic method of spike propagation that can significantly reduce the number of memory accesses required for evaluation of a spiking neural network, thus saving both time and energy.
The specific contributions of this paper are as follows:
\begin{itemize}
\item We show that by interpreting synaptic weights as probabilities, it is possible to implement spiking networks, and show that as the number of timesteps in the computation increases, the behaviour converges to that of the original (deterministically evaluated) network.
\item By altering the way the weights are stored and indexed, we achieve significant decrease in the number of memory accesses required to evaluate a network.
\item We present a hardware architecture that can be used as an accelerator on a System-on-Chip (SoC) platform, and can implement this model of computation efficiently.  Several optimizations on the architecture are presented that allow significant speedups over the software implementation.
\item The impact of this approach is quantified on well known benchmark circuits (MNIST and CIFAR-10).
\end{itemize}

The paper is organized as follows: we next present the motivation for the probabilistic interpretation of weights, and show through experiments that this can be implemented with minimal impact on accuracy of the network.
We quantify the reduction in memory accesses that would result even in a pure software implementation of the scheme.
Next, we present an architecture suitable for implementation as a hardware accelerator for an SoC, and study several variants of the probabilistic spike propagation that provide different accuracy \emph{vs} performance tradeoffs.
We then discuss the results in the context of previous approaches from the literature, and finally present our conclusions.

\section{Motivation}

A spiking neural network involves three kinds of operations, namely (a) injection, (b) generation and (c) propagation of spikes.
Spike injection involves injecting input spikes into the network, which would excite and initiate activity in the network. This is done either by generating spike trains from static inputs by following some probability distribution, or by input mechanisms that naturally generate spike trains.

Spike generation is the process of evaluating the neurons and their spiking in response to stimuli based on some mathematical model of the neuron. 
In this work, we have used the Integrate-and-Fire neuron model, but the approach is largely independent of the underlying model. 
The computational load of this part grows with the number of neurons.

Spike propagation decides which output synapses should be affected by a spike on a neuron. 
This requires identifying the recipients of the spikes and fetching the corresponding synaptic weights. 
The computational load for this grows with the number of synapses, making it the bottleneck of SNN evaluation. 

Consider a neuron $i$ with $N$ outgoing connections: every time it spikes, all of its outgoing connections must be updated. 
This requires accessing the $N$ synaptic weights and updating the state variables of the postsynaptic neurons. 
In the case of fully connected networks, or the fully connected layers of a deep network, the number of connections and weights can be very large, and the memory accesses here can dominate the overall performance of the network.
In the present work, we focus on these layers as they are the ones with the highest ratio of memory to computation.

\begin{figure}
    \begin{subfigure}[b]{\linewidth}
        \centering
        \includegraphics[width=0.7\linewidth]{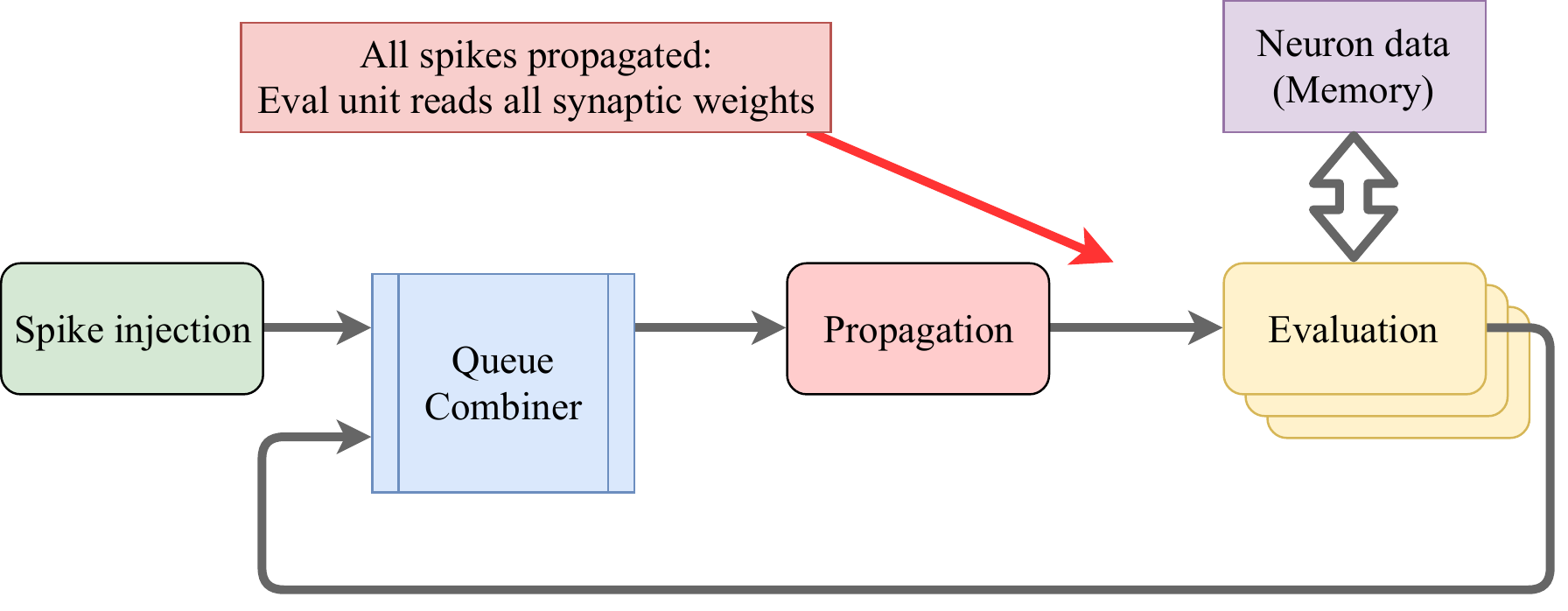}
        \caption{All synapses evaluated}
        \label{fig:q-arch-base}
    \end{subfigure}
    \begin{subfigure}[b]{\linewidth}
        \centering
        \includegraphics[width=0.7\linewidth]{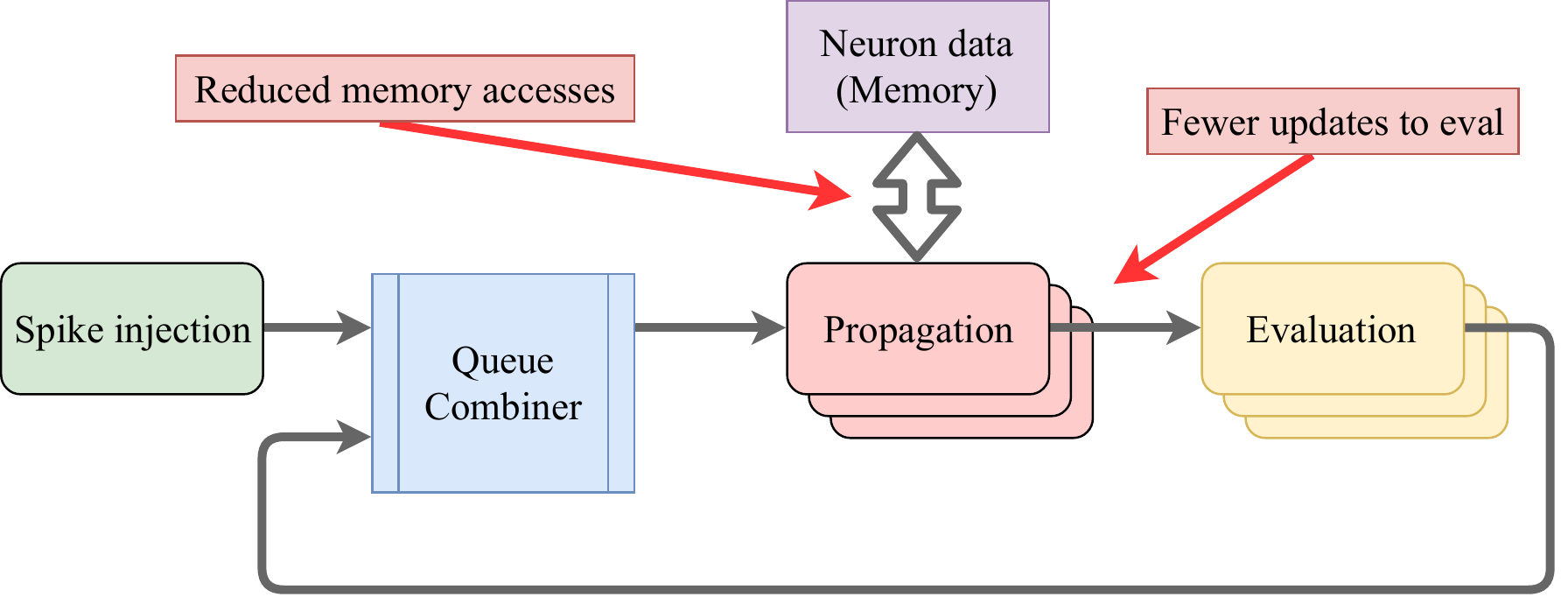}
        \caption{Probabilistic synapse evaluation}
        \label{fig:q-arch-prob}
    \end{subfigure}
    \caption{Schematic architectures for spike processing.}
\end{figure}

Fig.~\ref{fig:q-arch-base} shows a schematic architecture for this approach: the spike injection unit as well as the neuron evaluation unit feed into a \emph{queue} of spikes, that are in turn processed by the evaluation unit.  
This unit needs to read in the weights of all outgoing synapses for each spiking neuron, and update the target's membrane potential by an amount equal to the weight of the synapse.
In this architecture, almost all the work (and memory access) happens in the evaluation unit, and the propagation unit just passes generated spikes through to the queue for the next timestep.

Fig.~\ref{fig:q-arch-prob} shows an alternate view where the decision on which neurons are to be updated in the next timestep is made by the spike propagation unit, which decides whether a spike propagates across a synapse, while the evaluation unit makes the final decision on spiking and inserts entries into the queue for the next timestep.
In the next section, we will see how this change can be used to reduce memory accesses.

\subsection{Probabilistic spike propagation}
\label{sec:prob-spike-prop}
\begin{figure}[htbp]
    \centering
    \includegraphics[width=0.7\linewidth]{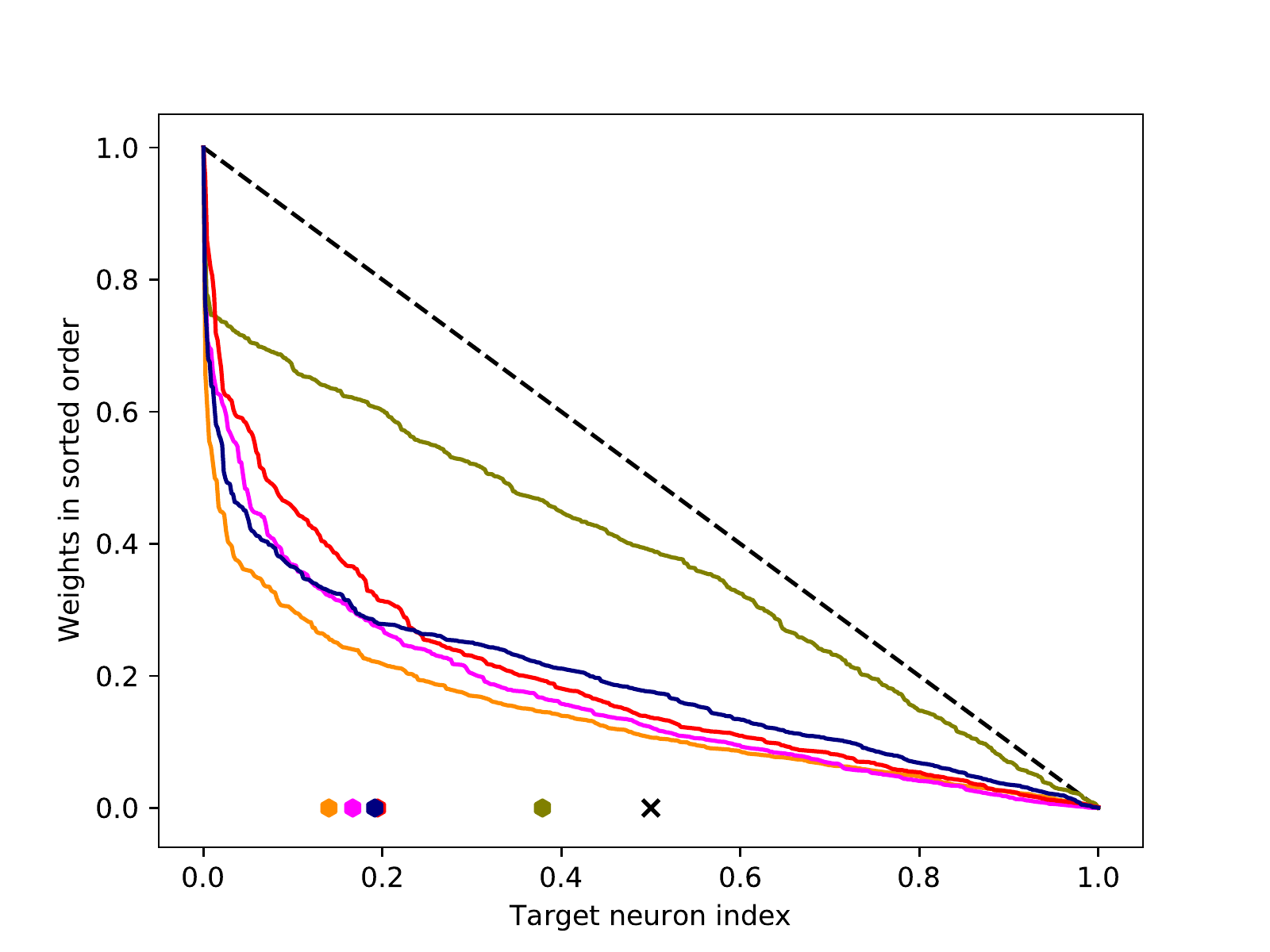}
    \caption{Weights of sample neurons in descending order, normalized against maximum outgoing weight for that neuron; $x$-axis is index of target neuron normalized by number of outgoing synapses.  Expected \emph{termination points} are marked as dots on the $x$-axis.} 
    \label{fig:term_point}
\end{figure}

The typical weight distribution for a neuron shows a few synapses with large weights, tapering down to a relatively large number of synapses of low weight. 
For example, fig.~\ref{fig:term_point} shows the outgoing synaptic weights of a few sample neurons in a network for the MNIST dataset.
Note that if the weights were uniformly distributed, we would expect this chart to be a straight (diagonal) line. 
For a weight distribution skewed towards the lower bound the curve would be below this diagonal.

One way to take advantage of this is to quantize the weights, and suppress those below a threshold.
However this impacts accuracy, and the resulting loss cannot be recovered.

We propose an alternate interpretation of a synaptic weight in terms of the probability of propagating a spike on that synapse.
By limiting the randomness to the propagation of spikes alone and leaving other aspects of the neuron model unchanged, we show that we can use weights from existing spiking networks without major changes.

In the deterministic approach, when neuron $i$ spikes, the applied weight for every postsynaptic neuron update is equal to the actual weight $w_{ij}$. 
For the integrate-and-fire neuron model, over a set of $N_i$ spikes, this will cause an increase in membrane potential of $N_i\times w_{ij}$ on neuron $j$.
If this exceeds the threshold voltage, a spike is produced, and the membrane potential is reset.
We hypothesize that it is enough that the temporal sum of the applied weights, across multiple spikes of $i$, for each neuron $j$ should approach $N_i\times w_{ij}$.
In other words, if we apply a weight $\hat{w}_{ij}$ with probability $p_{ij}$, then it is sufficient that
\begin{equation}
    N_i \times w_{ij} = N_i\times p_{ij}\times \hat{w}_{ij}
\end{equation}

This can be achieved by letting $p_{ij}=\frac{w_{ij}}{\wmaxi}$ and $\hat{w}_{ij}=\wmaxi$, where \wmaxi{} is the maximum weight for outgoing synapses of neuron $i$.  
When neuron $i$ spikes, we generate a random number $r$ uniformly distributed between $[0, \wmaxi]$, compare $r$ with $w_{ij}$ and update only those postsynaptic neurons $j$ for which the comparison succeeds, with an applied weight of \wmaxi{}.

Note that we process \emph{excitatory} and \emph{inhibitory} synapses separately as they correspond to positive and negative weights respectively. 
Both types are treated in the same way, but for the inhibitory synapses we let $p_{ij} = \frac{w_{ij}}{w_i^{min}}$ and $\hat{w}_{ij} = w_i^{min}$.

\begin{figure}
    \centering
    \includegraphics[width=0.6\linewidth]{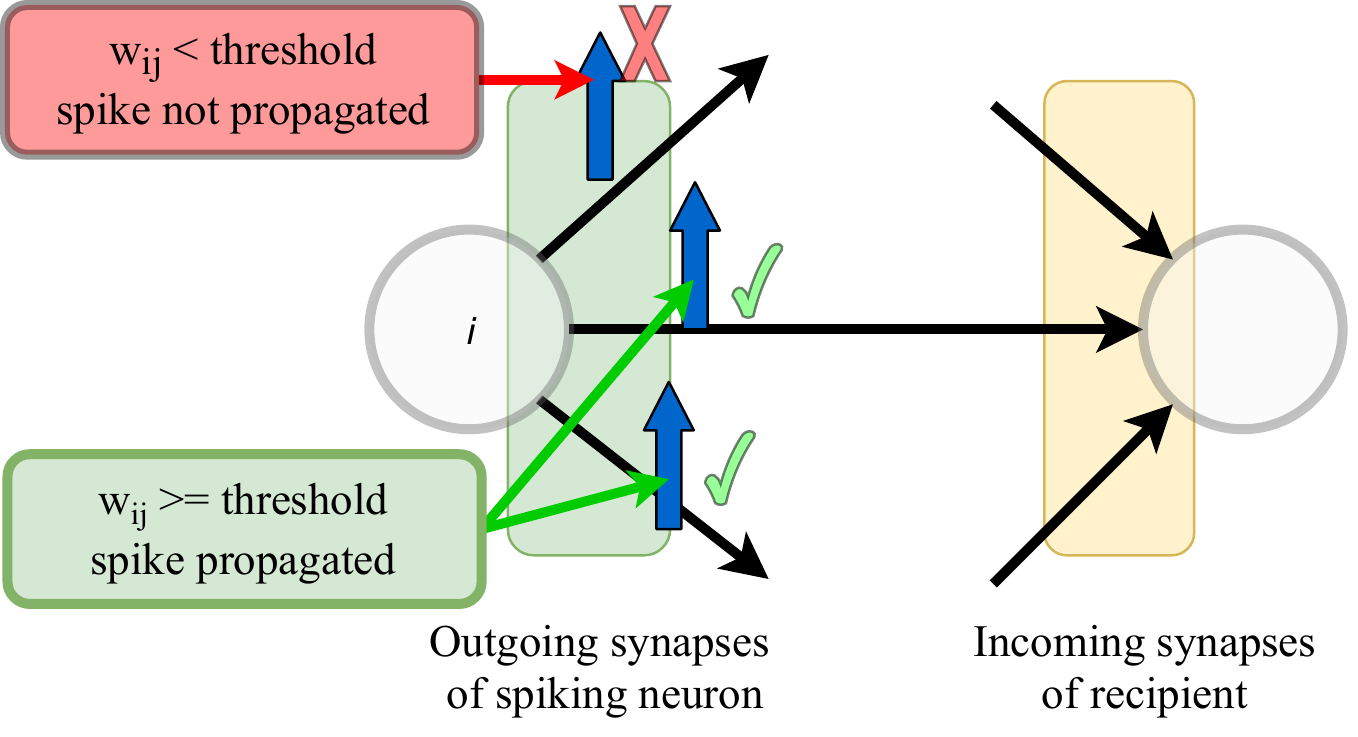}    
    \caption{Probabilistic spike propagation.}
    \label{fig:neuron-view}
\end{figure}

\subsection{Reducing memory accesses}
Probabilistic spike propagation can be leveraged to reduce memory fetches for outgoing synaptic weights by sorting the weight array. 
We generate a random value $r$ per spiking neuron, and propagate the spike to those targets with $w_{ij}>r$.
This requires reading the weight and index of the postsynaptic neuron, but as soon as the comparison fails for one synapse, all remaining synapses can be skipped.

Note that we need to store both the actual weight value and the index of the target neuron, potentially requiring twice the amount of memory, and twice the number of memory accesses.  
Therefore, for this approach to be useful, we will need to reduce both these requirements.

Assume the target neuron indices for the outgoing synapses of neuron $i$ are numbered from $1$ to \Nmaxi. 
Note that \Nmaxi could vary from one neuron to another -- this can account for differences between layers or even in the number of excitatory and inhibitory connections a given neuron has.
We define the \emph{termination point} of the weight update for neuron $i$ as the smallest index $t\in [1,\Nmaxi]$ for which $sorted\_weight_i[t] < r$.
It is clear that we would like the average value of $t$ over all neurons and all timesteps to be small in order to save on memory accesses. The expectation of this $t$ approaches $\sum_j w_{ij}/\wmaxi$.
In particular, if $E[t] < \Nmaxi/2$, then even if we access weights and indices separately to propagate spikes, we still need fewer memory accesses than the original deterministic approach.

This condition is satisfied even for the simple case that the weight distribution for a given neuron is uniform.
Since $r$ is drawn from uniform distribution between $0$ and \wmaxi{}, across multiple spikes on $i$ the expected value ($E[r]$) would approach $\wmaxi/2$.
On the other hand, for the type of distribution discussed in sec.~\ref{sec:prob-spike-prop}, the termination point would shift to smaller values, as can be seen in fig.~\ref{fig:term_point}.
Here, most neurons have termination points much lower than $\Nmaxi/2$, which implies that the probabilistic method would need fewer total memory accesses than the deterministic approach.

\subsection{Figure of merit}
We introduce the term \emph{Memory accesses per spike} (MAPS) to quantify and compare different implementations of a spiking network. 
When a neuron spikes, we measure the number of memory accesses (synaptic weights or indices) that are required to \emph{process} the spike, or update neurons affected by this spike.

For example, if the output of a given neuron in a SNN is connected to $N$ other neurons, then in the baseline deterministic approach, we would need to read in the weights for all $N$ synapses, and perform updates on all $N$ of the output neurons.
However, with the probabilistic approach, it may be possible to process only a subset of these for a given spike.
This will result in a lower MAPS.

Note: it is quite possible that the neuron indices and the synaptic weights require different numbers of bits for storage, which would also impact the energy consumed for reading one of these values.
This would vary significantly from one network to another, and there are also known techniques to try and further compress the storage and bandwidth requirements.
However, our purpose in the study here is to quantify the savings in memory accesses themselves, and we ignore differences in numbers of bits.

\section{Implementation Issues}
\label{sec:implementation}

We now consider how the probabilistic approach can be implemented, both on pure software platforms, as well as on custom hardware.  

\begin{algorithm}
    \DontPrintSemicolon
    \caption{Scan-based termination}
    \label{alg:swtrivial}
        \KwIn{$i$, $sorted\_weight$, $sorted\_index$} 
        $r\gets UniformRandom(0, \wmaxi)$ \;
        $j \gets 1$ \;
        \While{$sorted\_weight[j] \geq r$}{
            $UpdateNeuron(sorted\_index[j], \wmaxi)$ \;
            $j$\texttt{++} \;
        }
\end{algorithm}

Alg.~\ref{alg:swtrivial} implements the probabilistic approach by scanning weights until the value drops below the randomly generated threshold.  
This approach has the disadvantage that both the $sorted\_weight[i]$ and $sorted\_index[i]$ need to be read to process each outgoing synapse.

\begin{algorithm}
    \DontPrintSemicolon
    \caption{Termination with binary search}
    \label{alg:swbinsearch}
        \KwIn{$i$, $sorted\_weight$, $sorted\_index$} 
        $r\gets UniformRandom(0, \wmaxi)$ \;
        $termpt \gets BinarySearch(sorted\_weights, r)$ \;
        \For{$j \in [1, termpt]$}{
            $UpdateNeuron(sorted\_index[j], \wmaxi)$ \;
        }
\end{algorithm}

Alg.~\ref{alg:swbinsearch} modifies this to determine the termination point by running a binary search on the sorted weight array.
This can be done much faster ($O(\log \Nmaxi)$ comparisons) than the scanning based method (which will end up performing $termpt$ comparisons), and once the termination point has been determined, only the neuron indices ($sorted\_index$) need to be read.

\subsection{Hardware Architecture}
\label{sec:architecture}
\begin{figure}[htbp]
    \centering
    \includegraphics[width=0.7\linewidth]{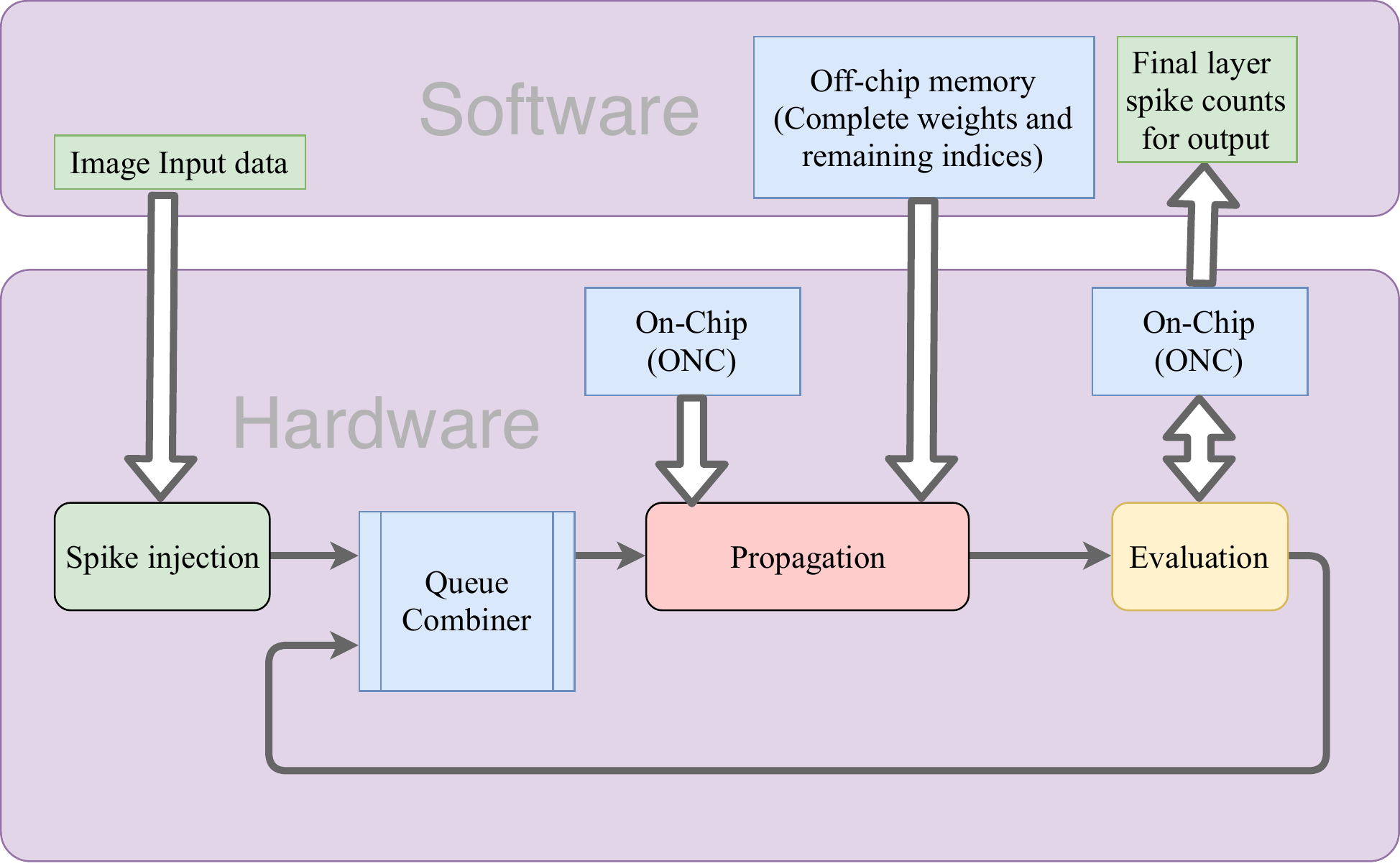}
    \caption{Hardware accelerator architecture}
    \label{fig:hw-arch}
\end{figure}

Though the proposed approach already shows advantages in a software implementation, it is possible to extract even more benefit from a hardware architecture that is able to exploit the memory access patterns appropriately.
Rather than designing a full architecture for spiking network processing, we designed a hardware accelerator in an SoC system built around the ARM processor core of a Xilinx Zynq FPGA.
A schematic of the architecture is shown in fig.~\ref{fig:hw-arch}.

The main tradeoff we are investigating is to store some weights on-chip in exchange for much lower off-chip access.  
FPGAs are well suited to this kind of architecture, as they have built-in RAM blocks that can be used for such storage, whereas this could be quite expensive in an ASIC.
So even though the core ideas here would apply to an ASIC as well, the benefits are easiest to realize on an FPGA platform where the approach is implemented as a hardware co-processor for the system CPU.

We now examine some variants of the probabilistic method, and how they would impact the resulting architecture and performance.

\subsubsection*{Binary search}
The basic binary search based method allows us to quickly compute a termination point and then use a simple iteration through indices to update the outgoing synapses.
While this is a good fit for software, in hardware the random access to weights at different indices that is required for binary search can be a problem, since off-chip memory accesses in hardware are usually best performed in a sequential burst.

\subsubsection*{Random index}
Since the sole purpose of the binary search is to compute a termination point for alg.~\ref{alg:swbinsearch}, we examine alternative approaches to solve this problem. 
The most obvious solution is to just generate $termpt$ randomly, rather than generating $r$ and mapping it to $termpt$.
This completely eliminates the need to access weights to decide how many synapses are to be updated, but the drawback is that it also loses any information contained in the weight distribution.
As a result, the random index based termination method performs poorly on networks where the majority of weights follow distributions other than uniform.

\subsubsection*{Weight transformed indexing}
Another possibility is to generate a random index $x\in [1,\Nmaxi]$, and use a function $fw_i(x)$ to translate that into a termination point:
\begin{equation}
    termpt = fw_i(x)
\end{equation}
With this formulation, it can be shown that if the normalized weight distribution (weights normalized against \wmaxi as a function of neuron index normalized against \Nmaxi) satisfies the equation
\begin{equation}
    f(f(x)) = x 
    \label{eq:selfinverse}
\end{equation}
then we can use a transform function of the form 
\begin{equation}
    termpt = w_i[x]\times \frac{\Nmaxi}{\wmaxi}
\end{equation}
It turns out that eq.~\ref{eq:selfinverse} is reasonably closely satisfied for several practical sorted weight distributions, similar to those seen in fig.~\ref{fig:term_point}.
For example, a linearly decaying sorted weight distribution or a hyperbolic distribution will both satisfy this condition.

This implies that we can compute the $termpt$ using a single weight lookup: generate a random index $x$, look up $w_i[x]$ and transform it to $termpt = w_i[x]\times \Nmaxi/\wmaxi$. 
Although this requires one random access to the weight memory, this is considerably less expensive than the binary search, and in most cases is found to perform better than the pure random $termpt$ method described previously.

For a hardware implementation, we need to store both sorted weights and sorted indices in memory (which could be off-chip high density DRAM for example), but only the sorted indices will need to be read in large quantities: weights will only be accessed to map the $termpt$.

\subsubsection*{Piecewise linear approximation}
The final approach we consider further trades off additional storage for accuracy.
Here, we use the observation from fig.~\ref{fig:term_point} that several of the sorted weight distributions seem to show a pattern of piecewise linear segments.
This means that if we can use some additional storage to keep track of indices where the weight distribution changes slope, we could get better accuracy, and closer fidelity to the original sorted weight distribution, while not incurring the full cost of binary search or other methods.

\subsection{Experimental results}
\label{sec:results}
The proposed approaches were validated on well-known benchmark problems (MNIST and CIFAR-10), using multiple networks that were trained as ANNs and converted to SNNs using the methods in \cite{Diehl2015Fast-classifyingBalancing}.
The networks and the different probabilistic approaches are listed in tables \ref{tab:benchmarks} and \ref{tab:termpt}.

\begin{table}
    \centering
    \setlength{\tabcolsep}{8pt}
    \begin{tabular}{ccc}
        \hline
        \textbf{Network} & \textbf{Architecture} & \textbf{Layers} \\
        \hline
        MNIST1  & Fully connected & 784-1200-1200-10 \\
        MNIST2  & Convolutional   & 5x5x32c-2x2p-5x5x64c-\\
                &                 & -2x2p-2048-10\\
        CIFAR10 & Convolutional   & 3x3x64c-2x2p-3x3x128c-\\
                &                 & -2x2p-3x3x256c-2x2p-1024-10 \\
        \hline
    \end{tabular}
    \caption{Benchmark networks: the \emph{Layers} indicate the number of neurons in each layer, with \emph{c} for convolutional, \emph{p} for average-pooling, and others fully connected.}
    \label{tab:benchmarks}
\end{table}

\begin{table}
    \centering
\begin{tabular}{cc}
    \hline
    \textbf{Technique} & \textbf{Termination point} \\
    \hline
    DET & Deterministic (baseline) \\
    BS & Binary search \\
    RI & Random index chosen \\
    TR & Weight transform function \\
    PWL & Piecewise linear with 5 segments \\
    \hline
\end{tabular}
\caption{Techniques to determine the termination point for the probabilistic approach.  DET is the deterministic baseline where all neurons are updated.}
\label{tab:termpt}
\end{table}

\begin{table}[t]
    %
    \centering
\begin{tabular}{|c|c|c|c|c|c|c|c|}
\hline
\multirow{2}{*}{ \bf{Network} } & \multirow{2}{*}{ \bf{Prop} } & \multirow{2}{*}{ \bf{Avg MAPS }} & \multicolumn{5}{c|}{\bf Fraction of weights on-chip} \\ \cline{4-8}
& & & \bf 0 & \bf 0.2 & \bf 0.4 & \bf 0.6 & \bf 0.8 \\ \hline
\multirow{2}{*}{ \bf MNIST1 } &  DET & 1200 & 1 & 0.8 & 0.6 & 0.4 & 0.2 \\ \cline{2-8}
& PWL & 240 & 0.20 & 0.11 & 0.06 & 0.03 & 0.01 \\ \hline
\multirow{2}{*}{ \bf MNIST2 } & DET & 2048 & 1 & 0.9 & 0.8 & 0.7 & 0.6 \\ \cline{2-8}

& PWL & 684 & 0.33 & 0.21 & 0.12 & 0.05 & 0.01 \\ \hline
\multirow{2}{*}{ \bf CIFAR10 } & DET & 1024 & 1 & 0.9 & 0.8 & 0.7 & 0.6 \\ \cline{2-8}
& PWL & 250 & 0.24 & 0.25 & 0.08 & 0.00 & 0.00 \\ \hline
\end{tabular}
\caption{Average memory accesses per spike (MAPS) when a fraction of the weights/indices are stored on-chip (MAPS normalized against the baseline deterministic value)}
\label{tab:memtable}
\end{table}

We conducted experiments on the different approaches to quantify the loss in accuracy compared to the deterministic approach, and to estimate how many more timesteps are required by the probabilistic method to reach the same accuracy as the deterministic approach.
The piecewise linear approximation seems to be the best compromise between hardware complexity (storage required for the piecewise linear breakpoints) and the accuracy.

Table \ref{tab:memtable} summarizes the reduction in off-chip memory access that can be obtained through use of the probabilistic methods.
As off-chip memory typically consumes considerably more energy (and has higher latency)\cite{Han2016EIE:Network} than on-chip memory, we would like to move as many of the weight accesses to on-chip storage as possible.
Assuming we have a fixed amount of memory is available on-chip, we would like to know which weights should be stored on-chip to get the best benefit.

The deterministic approach has to read all outgoing synaptic weights in any case, so cannot benefit from this, but the probabilistic methods can gain considerably here. 
As we can see from the table, for the benchmark networks considered, if we can store around 40\% of the weights on-chip, the number of off-chip accesses drops to very low levels.

Most importantly, we can trade off the on-chip storage for reduction in off-chip accesses much more effectively than when doing deterministic evaluation.

\begin{table}[t]
\centering
\begin{tabular}{|c|c|c|c|c|c|c|}
\hline
\textbf{Network}  & \textbf{Steps} & \textbf{DET} & \textbf{BS} & \textbf{RI} & \textbf{TR} &  \textbf{PWL} \\
\hline
\multirow{4}{*}{ \bf MNIST1} 
& 100 & 98.55 & 98.48 & 97.49 & 98.29 &  98.43 \\
& 200 & 98.53 & 98.5 & 97.51 & 98.29 &  98.47 \\
& 300 & 98.54 & 98.5 & 97.5 & 98.29 &  98.49 \\
& 1000 & 98.56 & 98.52 & 97.5 & 98.29 &  98.51 \\
\hline
\multirow{4}{*}{ \bf MNIST2}
& 100 & 99.13 & 98.96 & 99.05 & 99.07 & 99 \\
& 200 & 99.17 & 99.05 & 99.13 & 99.11 & 99.1 \\
& 300 & 99.16 & 99.09 & 99.12 & 99.12 & 99.13 \\
& 1000 & 99.17 & 99.12 & 99.14 & 99.11 & 99.14 \\ \hline
\multirow{3}{*}{ \bf CIFAR10}
& 200 & 74.35 & 71.73 & 72.51 & 73.13 & 72.4 \\
& 400 & 76.05 & 74.79 & 75.63 & 75.14 & 74.92 \\
& 600 & 76.44 & 75.4 & 76.87 & 75.62 & 75.74 \\
\hline
\end{tabular}
\caption{Accuracy of the probabilistic methods converges towards that of deterministic as the number of simulation steps increases.}
\label{tab:sim-accuracy}
\end{table}

Table \ref{tab:sim-accuracy} shows how the accuracy of the probabilistic methods converges to that of the deterministic method as we run for more simulation steps. 
Even though this looks like it may be a negative point for the probabilistic approach, the fact that the number of memory accesses has been reduced disproportionately (table \ref{tab:memtable}) means that the total memory accesses are in fact lower than the deterministic approach, even after accounting for the additional simulation time.

\begin{figure}
    \centering
    \includegraphics[width=0.7\linewidth, angle=0]{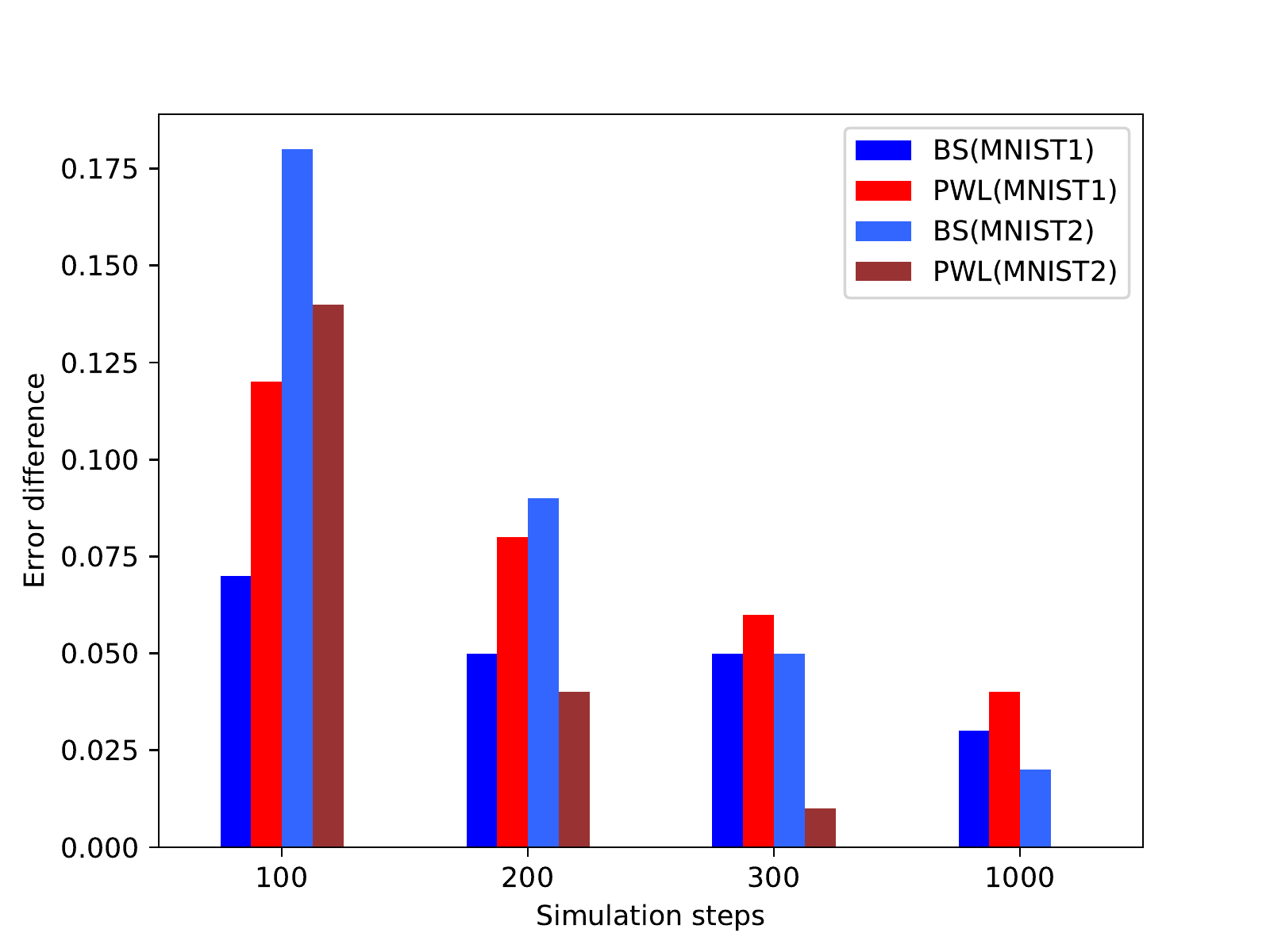}
    \caption{Difference in error percentage over increasing run times between deterministic and probabilistic spike propagation for the MNIST1 and MNIST2 networks.}
    \label{fig:errdiff}
\end{figure}

Fig. \ref{fig:errdiff} highlights the accuracy convergence: for the two variants of MNIST, we see how the difference between the deterministic and two probabilistic methods decreases with more timesteps.    

\section{Discussion}
\label{sec:discussion}

The focus of this work is on reducing memory accesses in implementation of spiking neural networks.
We have achieved this by introducing a probabilistic approach to spike propagation, that allows weights of known pre-trained spiking networks to be used, without requiring retraining or restructuring the network.
We now relate this to previously published works in this area, and bring out the differences in our approach.

\subsubsection*{Use of pre-trained networks}
Training spiking neural networks is not as well developed as training conventional artificial neural networks (ANNs), and the level of accuracy obtained using methods such as spike timing dependent plasticity (STDP) \cite{Diehl2015Fast-classifyingBalancing} are generally lower than the state of the art ANNs.
Previous research (eg. \cite{Diehl2015Fast-classifyingBalancing,Rueckauer2017ConversionClassification}) has shown how weights obtained by training an ANN can be converted for use with a spiking network model, with minimal impact on accuracy. 
A problem with this approach is that the resulting spiking networks are often quite complex.
In particular, the expected benefits of spiking networks: less complex compute units, fewer and sparser weights and activations etc. may not be obtained by this method of conversion.
However, since they provide high accuracy spiking networks, they are often used as a reference to compare a spiking network against an ANN.

In this paper, we are not concerned with how the spiking network is obtained, only whether we can interpret the synaptic weight as probability of transmission of a spike.
We try to show how the benchmark circuit's accuracy can be retained at a high level with the conversion to our probabilistic approach, and not to try and improve on it in any way.
In fact, we do not attempt to compare against the original ANN, and are only interested in retaining fidelity to the converted SNN, and comparing the number of memory accesses that would be required for implementing it.

\subsubsection*{Custom hardware architectures}
There have been several custom hardware accelerators designed expressly to implement spiking networks, such as \cite{Akopyan2015TrueNorth:Chip,Davies2018Loihi:Learning,Neil2014MinitaurAccelerator,Cheung2016NeuroFlow:Processors,Smaragdos2017BrainFrame:Simulations} \emph{etc.}.
These accelerators usually aim to minimize power or energy, for which they introduce certain restrictions on the type of spiking network that can be realized effectively.
For example, \cite{Davies2018Loihi:Learning} specifically takes up a custom benchmark example of the LASSO algorithm to demonstrate the effectiveness of the spiking architecture when compared to a software implementation, and similar custom benchmarks are considered by \cite{Akopyan2015TrueNorth:Chip}.
This is not in itself a shortcoming of those architectures -- it could in fact point to a problem with how spiking networks are currently being applied to problems they are not well suited to. 

The memory access reduction we consider in this work applies uniformly to any spiking network architecture, including the custom approaches mentioned above.
In fact, most of the work on the custom architectures uses techniques such as bitwidth reduction or weight truncation to reduce memory traffic.
Our approach is complementary to such techniques, but would require the weights and indices to be stored in sorted order. 
If this is possible for a given hardware architecture, then it is possible to apply our ideas to these custom hardware systems as well and further improve their performance.

\subsubsection*{Stochastic techniques}
Stochastic computation techniques apply randomness to the process of computation itself \cite{Shanbhag2010StochasticComputation}.  
Variants of this approach have been applied to spiking neural networks (eg. \cite{Smithson2016StochasticNetworks,Ahmed2016ProbabilisticProcessor,Rossello2012Probabilistic-basedImplementation}).
These are mostly orthogonal to the ideas we discuss, since a different (stochastic) hardware architecture for individual compute units can also be incorporated into our approach.

\cite{Ahmed2016ProbabilisticProcessor} considers the probabilistic model of the neuron itself, but here the neuron is modified so that the spiking behavior itself is stochastic. 
We want to use pre-existing neuron models without changing the intrinsic spiking behavior, so only the spike propagation is made random in such a way as to exploit the time averaging.

Bayesian spiking neurons \cite{Deneve2008BayesianInference,PaulinBayesianNeurons} apply probabilistic techniques for the neuron models themselves.  
These and similar probabilistic neuron models such as \cite{Kasabov2010ToModel} focus on improving the functionality and scope of neuron models, rather than hardware implementation.

\subsubsection*{Approximate and Emerging technologies}
Approximate computing is well known in the area of signal processing and neural network hardware, but has seen limited application to spiking networks.
One example is \cite{Sen2017ApproximateNetworks}, where neurons are progressively trimmed from evaluation as time progresses.
Again, our approach is orthogonal to this, and could be used to further reduce computations even for those neurons that are being evaluated.

Finally, there are approaches that rely on the use of new and emerging technologies, such as spin-based computing (\cite{Srinivasan2017MagneticNetwork}). 
These are out of the scope of the present work as they completely change the way in which networks are implemented, and everything from the neuron model to the training is different.

\section{Conclusions}
\label{sec:conclusions}
Repeated accesses to synaptic weights forms the main bottleneck in the evaluation of spiking neural networks, especially in fully connected layers involving large numbers of weights.
The probabilistic approach to spike propagation presented in this paper can result in significant savings in the number of memory accesses required to evaluate a spiking neural network. 
The approach can be applied to a pre-trained spiking network without imposing restrictions on the type of network or the weights, and without requiring retraining.
This is made possible by the observation that the long-term average weight applied by this probabilistic approach over a number of timesteps converges to the actual synaptic weight that should have been applied in a deterministic approach.

Experiments on benchmark circuits show that the proposed approach is able to achieve equivalent results to a deterministic spiking network, given enough timesteps.
Even though the number of timesteps may be more, the probabilistic approach is able to achieve the same level of accuracy as a deterministic approach using fewer total memory accesses, which would translate directly into a lower total energy of computation.
For the benchmark circuits considered, by storing just 40\% of the weights from the fully connected layer on chip, we can reduce the number of off-chip memory accesses by close to 90\%.

\bibliographystyle{plain}
\bibliography{Mendeley_Abinand_edited}

\begin{thebibliography}{10}

\bibitem{Ahmed2016ProbabilisticProcessor}
K.~Ahmed et~al.
\newblock {Probabilistic inference using stochastic spiking neural networks on
  a neurosynaptic processor}.
\newblock In {\em IJCNN '16}, pages 4286--4293. IEEE, 7 2016.

\bibitem{Akopyan2015TrueNorth:Chip}
F.~Akopyan et~al.
\newblock {TrueNorth: Design and Tool Flow of a 65 mW 1 Million Neuron
  Programmable Neurosynaptic Chip}.
\newblock {\em IEEE Trans. on CAD}, 2015.

\bibitem{Cheung2016NeuroFlow:Processors}
K.~Cheung et~al.
\newblock {NeuroFlow: A General Purpose Spiking Neural Network Simulation
  Platform using Customizable Processors}.
\newblock {\em Frontiers in Neuroscience}, 9:516, 1 2016.

\bibitem{Davies2018Loihi:Learning}
M.~Davies et~al.
\newblock {Loihi: A Neuromorphic Manycore Processor with On-Chip Learning}.
\newblock {\em IEEE Micro}, 38(1):82--99, 1 2018.

\bibitem{Deneve2008BayesianInference}
S.~Deneve.
\newblock {Bayesian Spiking Neurons I: Inference}.
\newblock {\em Neural Computation}, 20(1):91--117, 1 2008.

\bibitem{Diehl2015Fast-classifyingBalancing}
P.~Diehl et~al.
\newblock {Fast-classifying, high-accuracy spiking deep networks through weight
  and threshold balancing}.
\newblock In {\em Proc. Intl. Joint Conf. on Neural Networks}, 2015.

\bibitem{Furber2014TheProject}
S.~Furber et~al.
\newblock {The SpiNNaker Project}.
\newblock {\em Proceedings of the IEEE}, 102(5):652--665, 5 2014.

\bibitem{Han2016EIE:Network}
S.~Han et~al.
\newblock {EIE: Efficient Inference Engine on Compressed Deep Neural Network}.
\newblock In {\em Proc. ISCA '16}, volume~44, pages 243--254. IEEE, 6 2016.

\bibitem{Kasabov2010ToModel}
N.~Kasabov.
\newblock {To spike or not to spike: A probabilistic spiking neuron model}.
\newblock {\em Neural Networks}, 23(1):16--19, 1 2010.

\bibitem{Maass1997NetworksModels}
W.~Maass.
\newblock {Networks of spiking neurons: The third generation of neural network
  models}.
\newblock {\em Neural Networks}, 10(9):1659--1671, 12 1997.

\bibitem{Neil2014MinitaurAccelerator}
D.~Neil and S.-C. Liu.
\newblock {Minitaur, an Event-Driven FPGA-Based Spiking Network Accelerator}.
\newblock {\em IEEE Trans. on VLSI}, 22(12):2621--2628, 12 2014.

\bibitem{PaulinBayesianNeurons}
M.~G. Paulin and A.~Van~Schaik.
\newblock {Bayesian Inference with Spiking Neurons}.
\newblock In {\em arXiv: 1406.5115}, 2014.

\bibitem{Rossello2012Probabilistic-basedImplementation}
J.~L. Rossello et~al.
\newblock {Probabilistic-based neural network implementation}.
\newblock In {\em The 2012 International Joint Conference on Neural Networks
  (IJCNN)}, pages 1--7. IEEE, 6 2012.

\bibitem{Rueckauer2017ConversionClassification}
B.~Rueckauer et~al.
\newblock {Conversion of Continuous-Valued Deep Networks to Efficient
  Event-Driven Networks for Image Classification}.
\newblock {\em Frontiers in Neuroscience}, 11:682, 12 2017.

\bibitem{Sen2017ApproximateNetworks}
S.~Sen et~al.
\newblock {Approximate computing for spiking neural networks}.
\newblock In {\em Design, Automation {\&} Test in Europe Conference {\&}
  Exhibition (DATE), 2017}, pages 193--198. IEEE, 3 2017.

\bibitem{Seung2003LearningTransmission}
H.~Seung.
\newblock {Learning in Spiking Neural Networks by Reinforcement of Stochastic
  Synaptic Transmission}.
\newblock {\em Neuron}, 40(6):1063--1073, 12 2003.

\bibitem{Shanbhag2010StochasticComputation}
N.~R. Shanbhag et~al.
\newblock {Stochastic computation}.
\newblock In {\em Proc. DAC '10}, page 859, New York, New York, USA, 2010. ACM
  Press.

\bibitem{Smaragdos2017BrainFrame:Simulations}
G.~Smaragdos et~al.
\newblock {BrainFrame: a node-level heterogeneous accelerator platform for
  neuron simulations}.
\newblock {\em J. of Neural Engg.}, 14(6):066008, 12 2017.

\bibitem{Smithson2016StochasticNetworks}
S.~Smithson et~al.
\newblock {Stochastic Computing Can Improve Upon Digital Spiking Neural
  Networks}.
\newblock In {\em 2016 IEEE International Workshop on Signal Processing Systems
  (SiPS)}, pages 309--314. IEEE, 10 2016.

\bibitem{Srinivasan2017MagneticNetwork}
G.~Srinivasan et~al.
\newblock {Magnetic tunnel junction enabled all-spin stochastic spiking neural
  network}.
\newblock In {\em Proc. of DATE, 2017}, pages 530--535. IEEE, 3 2017.

\end{thebibliography}

\end{document}